\documentclass[10pt,twocolumn,letterpaper]{article}

\usepackage[pagenumbers]{wacv} 

\usepackage{graphicx}
\usepackage{amsmath}
\usepackage{amssymb}
\usepackage{booktabs}
\usepackage{multirow}
\usepackage{xcolor}
\usepackage{pifont}
\newcommand{\xmark}{\ding{55}}%
\newcommand*\samethanks[1][\value{footnote}]{\footnotemark[#1]}

\usepackage[pagebackref,breaklinks,colorlinks]{hyperref}

\usepackage[capitalize]{cleveref}
\crefname{section}{Sec.}{Secs.}
\Crefname{section}{Section}{Sections}
\Crefname{table}{Table}{Tables}
\crefname{table}{Tab.}{Tabs.}

\def\wacvPaperID{1128} 
\def\confName{WACV}
\def\confYear{2024}

\begin{document}

\title{Efficient Feature Distillation for Zero-shot Annotation Object Detection}

\author{Zhuoming Liu\thanks{Equal contribution.}, Xuefeng Hu\samethanks[1], Ram Nevatia\thanks{Corresponding author} \\
University of Southern California\\
{\tt\small  \{liuzhuom, xuefengh, nevatia\}@usc.edu}
}

\maketitle

\begin{abstract}
We propose a new setting for detecting unseen objects called Zero-shot Annotation object Detection (ZAD).
It expands the zero-shot object detection setting by allowing the novel objects to exist in the training images 
and restricts the additional information the detector uses to novel category names.
Recently, to detect unseen objects, large-scale vision-language models (e.g., CLIP) are leveraged by different methods. 
The distillation-based methods have good overall performance but suffer from a long training schedule caused by two factors.
First, existing work creates distillation regions biased to the base categories, which limits the distillation of novel category information.
Second, directly using the raw feature from CLIP for distillation neglects the domain gap between the training data of CLIP and the detection datasets, which makes it difficult to learn the mapping from the image region to the vision-language feature space.
To solve these problems, we propose Efficient feature distillation for Zero-shot Annotation object Detection (EZAD). 
Firstly, EZAD adapts the CLIP's feature space to the target detection domain by re-normalizing CLIP;
Secondly, EZAD uses CLIP to generate distillation proposals with potential novel category names to avoid the distillation being overly biased toward the base categories.
Finally, EZAD takes advantage of semantic meaning for regression to further improve the model performance. 
As a result, EZAD outperforms the previous distillation-based methods in COCO by 4\% with a much shorter training schedule and achieves a 3\% improvement on the LVIS dataset. Our code is available at \url{https://github.com/dragonlzm/EZAD}
\end{abstract}

\section{Introduction}\label{sec:intro}

Object detection is a fundamental task in computer vision.  
Typically, an object detector is trained on a dataset with specific categories and can't be extended to new categories. 
To detect instances of categories not seen during training, researchers started studying Zero-Shot object Detection (ZSD) and Open-Vocabulary object Detection (OVD).
In both ZSD and OVD settings,  a set of "base" categories is provided with annotated training examples (bounding boxes), and the rest, called "novel" categories, are not provided with any annotated examples. In ZSD, the names of the novel categories are known at training time, but it is further posited\cite{bansal2018zero} that the training data should not have any instances of the novel categories; this is very restrictive and requires training data to be screened for novel categories which defeats the purpose of the user adding categories at will. OVD does not impose this restriction but assumes that the novel category names are unavailable during training and may be added freely at inference time. Also, to detect more novel categories, the training of OVD sometimes needs additional images or captions. 
\begin{figure}[!t]
   \centering
   \includegraphics[width=1\linewidth]{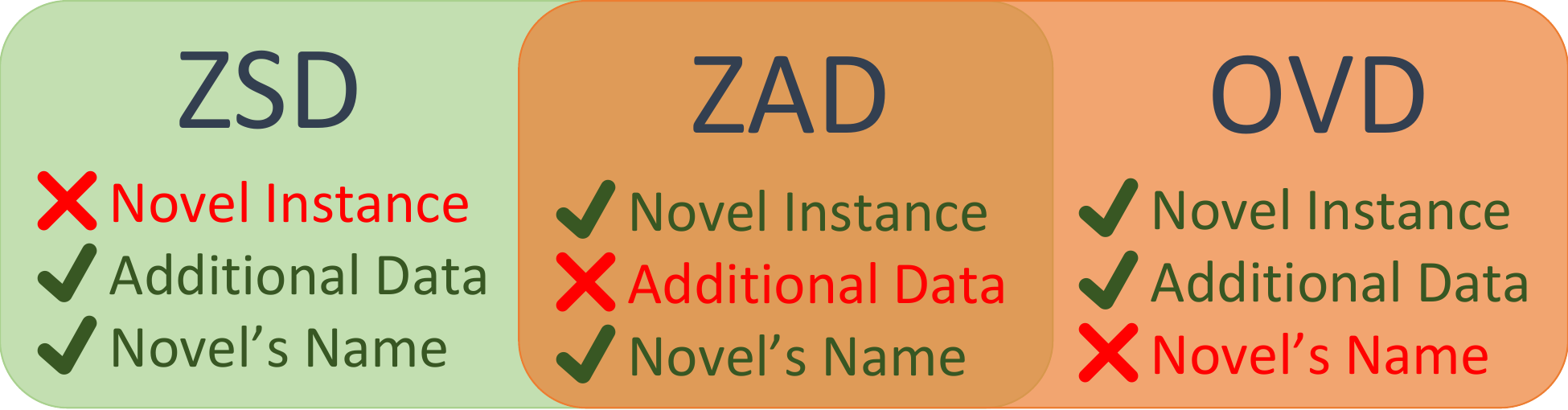} 
   \vskip -0.1in
   \caption{The difference between our zero-shot annotation object detection (ZAD), zero-shot object detection (ZSD), and open-vocabulary detection (OVD). Our ZAD can have novel instances in training images and novel category names, but no additional annotations are allowed.} 
   \vskip -0.1in
   \label{zadtest} 
   \vskip -0.2in
 \end{figure}

We propose an alternate setting that we call \textbf{Z}ero-shot \textbf{A}nnotation object \textbf{D}etection (\textbf{ZAD}) task, which allows the training data to contain unannotated novel category instances and a model to mine this information at training time if novel category names are available in advance. However, additional category names may be added at inference time, as in OVD. 
Thus, ZAD falls in between ZSD and OVD, as Fig~\ref{zadtest} shows. We believe ZAD is closer to a real-life setting: when no additional data or annotation is provided, a user wanting to add categories not considered at annotation time but exploiting the latent information as novel category names are added at the training time. 
  
To recognize novel instances when only their name is available, it is natural to consider a vision-language feature space.
In recent years, some large-scale vision-language classification models trained on millions of image-text pairs, for instance, CLIP\cite{radford2021learning}, ALIGN\cite{jia2021scaling} have become available.
Different solutions are proposed to leverage these models to enable a detector to detect novel objects.
Some researchers\cite{zareian2021open, gao2021towards, zhou2022detecting, zhong2022regionclip} use these models to generate pseudo-label and train their model with pseudo-label and the image-caption pair, while other trains a prompt\cite{feng2022promptdet, wu2023cora, kuo2022f} to turn the CLIP into a detector. 
However, all these methods need additional captions or images, which are unavailable in our ZAD setting.
In contrast, pure distillation-based methods\cite{gu2021open, zang2022open} learn a mapping from image regions to CLIP feature space by distillation for detecting the novel object, which is applicable when no additional data is provided.

To know how good the features are for distillation, we first apply CLIP to classify the instances in the COCO\cite{lin2014microsoft} dataset. 
We found that the classification accuracy (ACC) is only 46\% which is much lower than the ACC of the classifier in Faster R-CNN\cite{ren2015faster} (about 90\%). This indicates the domain gap between the training data of CLIP and the detection dataset, making the mapping from the image region to the vision-language feature space harder to learn. 
In addition, since the distillation is conducted on some specific image regions, how to select such a region is an important question.
ViLD\cite{gu2021open} trains an RPN with base category annotations and applies the RPN to the training image again to generate proposals as distillation regions.
As table~\ref{table:recall} shows, these proposals are biased toward the region with base categories instances, which limits the novel information obtained by the detector and harms the distillation efficiency. 

\begin{figure}[!t]
   \centering
   \includegraphics[width=1\linewidth]{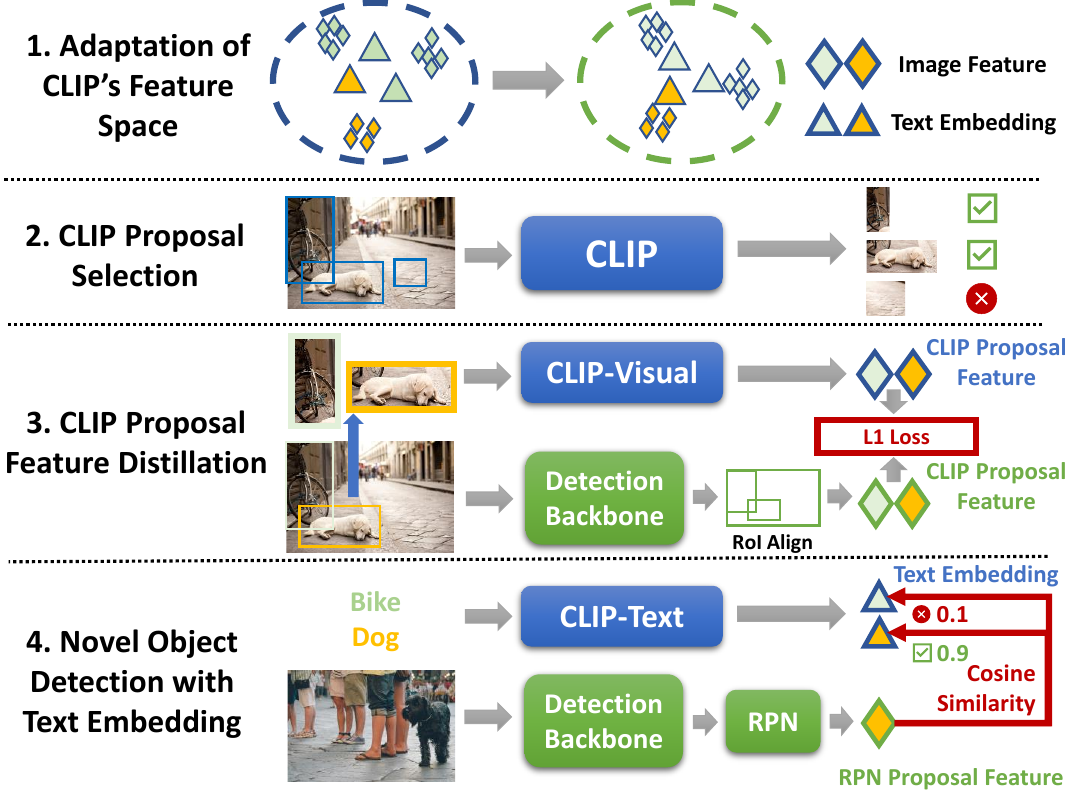} 
   \vskip -0.1in
   \caption{Overview of our method. The key contributions of EZAD are we bridge the domain gap and create better distillation regions (CLIP proposals).} 
   \label{clip_adaptaton_and_mapping} 
   \vskip -0.15in
 \end{figure}

To address these problems, we propose \textbf{E}fficient Feature Distillation for \textbf{Z}ero-shot \textbf{A}nnotation object \textbf{D}etection (\textbf{EZAD}).
For bridging the domain gap, we find that simply finetuning the LayerNorm layers in the CLIP with the base category instances significantly improves the ACC on both base and novel (Fig~\ref{clip_adaptaton_and_mapping}, 1).
For the distillation regions, we expect these regions could contain novel objects so that some novel category information can be introduced into the detector.
To make the best use of the only information we have in the ZAD setting, the name of novel categories, we decided to use CLIP to select these regions with the help of the novel category names. 
The selected regions are named CLIP Proposals, in which CLIP believes there is a novel category instance(Fig~\ref{clip_adaptaton_and_mapping}, 2).

After adapting the feature space and generating the CLIP Proposals, EZAD learns a mapping from the image regions to the vision-language feature space by distillation, which is achieved by minimizing the L1 loss between the features of the CLIP Proposals from the CLIP, and the one from our model (Fig~\ref{clip_adaptaton_and_mapping}, 3).
Once the model is trained, in all potential regions given by the RPN, EZAD recognizes the novel objects by using the novel category name's text embedding, which has a high cosine similarity score with the image feature of the novel objects (Fig~\ref{clip_adaptaton_and_mapping}, 4).
To further improve the model performance, we introduce a semantic-based regressor, which takes the text embedding as additional information for regression.

By only providing the name of the novel categories to EZAD, EZAD can outperform ViLD by 4\% in novel categories with a much shorter training schedule in COCO. On the LVIS dataset, our method achieves better performance than ViLD on both base and novel categories. This indicates that the adapted feature and CLIP Proposals benefit both distillation quality and training efficiency. 


  \begin{table}[!t]
    \centering
    \begin{tabular}{c|c|ccc}
      \hline 

       \hline
       Dataset & Eval On & AR@100 & AR@300 & AR@1000 \\
       \hline
       \multirow{2}{*}{COCO} & Base    & \textbf{56.69} & \textbf{61.45} & \textbf{64.32} \\
        & Novel   & 34.66 & 43.62 & 51.34 \\ \hline 
       \multirow{2}{*}{LVIS} & Base & \textbf{42.14} & \textbf{49.84} & \textbf{55.09}   \\
       & Novel   & 35.63          & 45.07          & 51.82 \\ 
      \hline     
    \end{tabular}
    \centering
    \vspace{-1 em} 
    \caption{The Recall of RPN on Novel and Base categories on COCO or LVIS training set. The RPN biases to the base category.}
    \vspace{-1.5em} 
    \label{table:recall}   
 \end{table}

\begin{figure*}[!t]
   \centering
   \includegraphics[width=1\linewidth]{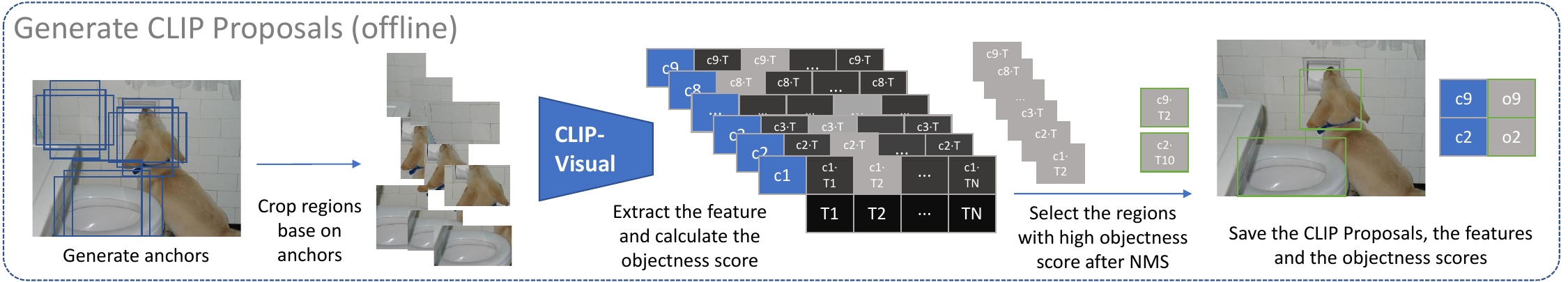} 
   \vskip -0.1in
   \caption{The pipeline of CLIP Proposals generation. The CLIP Proposals and their feature are generated before the detector training.} 
   \vskip -0.1in
   \label{clip_proposal} 
   \vskip -0.15in
 \end{figure*}

\section{Related Work}

\textbf{The Large-scale Pretraining} already exist in Vision (MOCO\cite{he2020momentum}, MOCOv2\cite{chen2020improved}, etc.) and Language (BERT\cite{devlin2018bert}, GPT\cite{brown2020language}, etc.) for a long time.
Recently, there is a trend to use free-form supervision (raw text) to train the vision model which has evolved to large-scale vision-language pre-trained models. 
For instance, the CLIP\cite{radford2021learning} and ALIGN\cite{jia2021scaling} are trained with a large-scale dataset with hundreds of millions of image-text pairs using contrastive learning.
GLIP\cite{li2022grounded} is pre-trained at an object-level with a large-scale grounding dataset.
These multi-modal models' feature space and their knowledge are useful and can be applied to many other tasks, such as zero-shot classification and zero-shot detection.
However, the training data of these models are usually noisy and there is a domain gap between these data and the datasets of downstream tasks.

\textbf{Domain Adaptation} is necessary when we apply a pre-trained model to other datasets.
In computer vision, the most common method for bridging the domain gap is to finetune the whole network on the new dataset or add one extra MLP layer at the end.
In the natural language processing community, Prompt tuning\cite{li2021prefix, lester2021power} surfaced as an important tool for domain adaptation in recent years.
Besides, the simple renormalization method is found to be effective by Perez\cite{perez2018film} and Lu\cite{lu2021pretrained}.
For adapting the large-scale multi-modal model, Kim et al.\cite{kim2022how} discuss the effectiveness of different ways in adapting the CLIP\cite{radford2021learning} to new classification datasets. As far as we know, no paper discusses bridging the domain gap between image-level and instance-level tasks.

\textbf{Zero-shot Detection object (ZSD) and Open Vocabulary object Detection (OVD)} aim to learn knowledge from the base categories and to generalize the knowledge to the novel categories, enabling the model to detect novel objects.
Bansal et al.\cite{bansal2018zero}, Zhu et al.\cite{zhu2020don}, Rahman et al.\cite{rahman2020improved}, and Xie et al.\cite{xie2022zero} focus on ZSD, in which novel instances cannot exist in the training image.
However, in most object detection datasets, not all the instances on the image will be annotated. 
This makes the ZSD setting too restricted.
Zhong et al.\cite{zhong2022regionclip}, Zhou et al.\cite{zhou2022detecting} and Zareian et al\cite{zareian2021open} pre-train their model with image-caption pair to learn the vision-language feature space for open vocabulary detection. 
Gao et al.\cite{gao2021towards}, Long et al.\cite{long2022p}, Wang et al.\cite{wang2023object} Zhao et al.\cite{zhao2022exploiting} train its model with pseudo-label.
Feng et al.\cite{feng2022promptdet}, Kuo et al.\cite{kuo2022f}, and Wu et al.\cite{wu2023cora} design a prompt to make use of CLIP for detection.
Ma et al.\cite{Ma_2022_CVPR} and Gu et al.\cite{gu2021open} distill visual features from CLIP, enabling object detectors to detect novel instances.
Wang et al.\cite{wang2023learning} propose CondHead which can be added to different OVD models to further improve their performance.
All these methods aim to solve the OVD problem, and most of them need additional training data except the distillation-based\cite{Ma_2022_CVPR, gu2021open, wang2023learning} and some prompt-based methods\cite{kuo2022f, wu2023cora}. 
The distillation-based and prompt-based methods can fit in with our ZAD setting, while the distillation-based method has better overall performance.

\begin{figure*}[!t]
   \centering
   \includegraphics[width=1\linewidth]{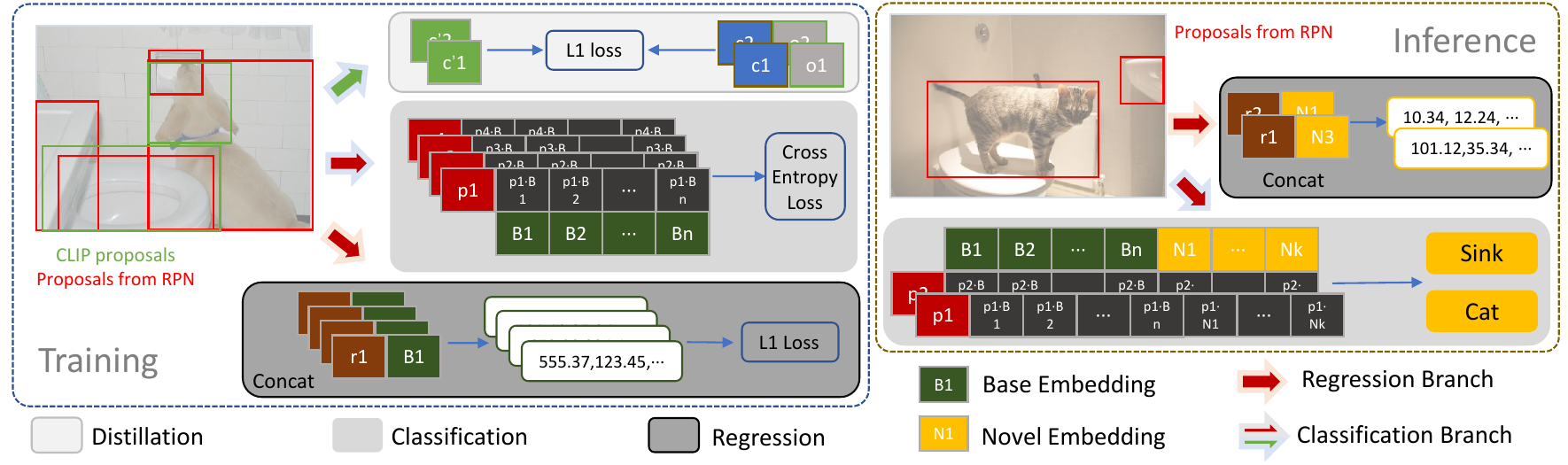} 
   \vskip -0.1in
   \caption{The training and inference of our model. The figure presents the model structures in the RoI head of the two-stage detector. We use the per CLIP Proposal distillation weight and semantic-based regressor to further improve the model performance.} 
   \vskip -0.1in
   \label{method_overview} 
   \vskip -0.1in
 \end{figure*}

\section{Method}
\textbf{Setting definition.} Our method aims to handle the \textbf{Z}ero-shot \textbf{A}nnotation object \textbf{D}etection (\textbf{ZAD}).
In ZAD, all categories are divided into two parts: the base categories $C_{b}$ and the novel categories $C_{n}$.
The name of the base and expected novel category are known before model training.
The training image may have novel category instances, but only the base category will be annotated.
No additional data will be available for the training except the training image and the base categories annotations.
The model is trained on a set of base classes and tested on both base and novel classes. 

\textbf{Method Overview.}
Inspired by the previous paper ViLD~\cite{gu2021open}, our proposed method, \textbf{E}fficient Feature Distillation for \textbf{Z}ero-shot \textbf{A}nnotation object \textbf{D}etection (\textbf{EZAD}), map the image feature to the CLIP's multi-modal feature space. 
The \textbf{\textit{mapping}} is learned by distilling the knowledge from CLIP in some selected \textbf{\textit{distillation regions}} of each training image. 
The knowledge distillation is conducted by optimizing the L1 loss between the \textbf{\textit{feature from CLIP}} and the feature from our model in the selected regions.
After learning the mapping, given the proposal from the RPN, our model can recognize and detect novel objects by using the text embedding of the novel categories.

In Section~\ref{finetune_feature}, we describe how to adapt the \textbf{\textit{feature from CLIP}} to the domain of detection datasets.
In Section~\ref{generate_clip_proposal}, we demonstrate how to select the \textbf{\textit{distillation regions}} on each image.
In Section~\ref{model_training}, we discuss our model structure and how to train our model and learn the \textbf{\textit{mapping}} with the adapted feature from the selected regions.

\subsection{Adapt Vision-language Model Feature} \label{finetune_feature}
To understand how good the image feature from CLIP is, we first apply the CLIP to classify the instances in the COCO~\cite{lin2014microsoft} dataset.
We first extract the feature for each instance from the vision encoder of the CLIP. 
The feature of the instance $i$ can be expressed as: $ins_{i} = V(Crop(I, $ $GT_{i(1.2x)}))$, where $V$ is the vision encoder of the CLIP and $Crop(I, GT_{i(1.2x)})$ means cropping the region from Image $I$ base on 1.2x enlarged GT bboxes $GT_{i(1.2x)}$.
We use the 1.2x enlarged bboxes since the enlarged bboxes help CLIP yield the best classification accuracy.
We present more details in the supplementary material.
We generate the text embedding for each COCO category from the text encoder of the CLIP.
We calculate the cosine similarity between the image feature and text embeddings and select the category with the highest cosine score as the predicted category.

We notice that when directly applying the CLIP to classify the COCO instances, the classification accuracy (ACC) is only about 50\% which is much lower than the ACC of the classifier in a well-trained detector, indicating that there is a huge distribution gap between the training data of the CLIP and detection datasets.
To bridge the gap, inspired by\cite{kim2022how}, we simply fine-tune the CLIP's layer of normalization layers by minimizing the cross-entropy loss, using all base categories instances in the detection dataset.
This simple method boosts the ACC on COCO to about 80\%.
Also, using the adapted CLIP feature for distillation helps improve detection results.

\subsection{Generate CLIP Proposals} \label{generate_clip_proposal}
To obtain useful information for the novel categories, we need to select some meaningful image regions for distillation.
we expect these regions to contain some novel category objects, 
and introduce information of the novel categories.
ViLD trains an RPN with base category annotations and uses the proposals from this RPN as the distillation regions.
However, since the RPN is trained to predict base categories, which makes the proposals bias toward the areas that contain the base categories and ignore the areas that potentially have the novel category instance.
Instead of using the RPN to determine where to distill, we decided to use CLIP~\cite{radford2021learning}.
Trained with 400 million image and text pairs collected from the internet, CLIP is trained to discriminate a large number of categories. 
Therefore, if the region contains a novel object, the CLIP should yield high confidence in it.
We name these regions CLIP Proposals.
Fig~\ref{clip_proposal} demonstrates how to generate CLIP Proposal.

To select image regions as the CLIP Proposals, we first generate anchors over the image $I$.
Then we crop the image based on the anchors and extract the feature for these anchors from CLIP's vision encoder.
We generate the text embeddings $T_{i}$ for a given dictionary from the CLIP's text encoder.
In addition to the information of the novel categories, we also need knowledge of the base categories. Thus, we use all category names as the dictionary.
We classify each anchor by calculating the cosine similarity score between its image feature and all text embeddings.
We use the score after the softmax of the predicted category as the objectness score $o_{i}$ of the anchor.
We finally select the anchors with high objectness scores after the Non-Maximum Suppression (NMS) as CLIP Proposals. 

All the CLIP Proposals $C_{i}$ and their feature from CLIP $c_{i}$ are generated offline.
The $c_{i}$ can be expressed like this: $V(Crop(I, C_{i}))$, where $V$ and $Crop$  are the same as those in $ins_{i}$ definition.
We also add 1.2x enlarged base categories' GT bbox as part of the CLIP Proposals.
Our experiments show that even though these CLIP Proposals are noisy, they are still meaningful regions for distillation and help our detector perform better on novel categories.

\subsection{Model Structure} \label{model_training}
Compared with the traditional two-stage detector, our model structure has three main differences: CLIP Proposals' feature distillation, cosine-similarity-based classifier, and semantic-based regressor. 
The model structure overview is shown in Fig \ref{method_overview}.

\textbf{Proposals' Feature Distillation.}
To obtain knowledge from CLIP and map the image feature from our model into the CLIP's feature space, we distill the knowledge from CLIP by minimizing the L1 loss between the CLIP Proposals' feature from the CLIP's vision encoder $c_{i}$ and the one from our model $c^{'}_{i}$. 
The $c^{'}_{i}$ can be expressed as $Conv_{c}(Align(Bbone(I), C_{i}))$, where $Bbone(I)$ means getting the feature map by passing the image $I$ through the backbone $Bbone$, and $Align$ means doing the RoIAlign base on the CLIP Proposal $C_{i}$ on the feature map. 
$Conv_{c}$ means passing the feature after the RoIAlign through the convolution and linear layers in the classification branch.

In the CLIP Proposal generation, we know the objectness score of each proposal.
For the proposal with a higher objectness score, it has a higher probability of having an object in it.
Therefore, we should assign a higher weight to these proposals. 
We directly use the objectness score as the weight, and the distillation loss is formulated like this:
\begin{equation}
   \label{eq:gradsimi}
   \begin{aligned}
   L_{dist} = \frac1M\sum_{i=1}^Mo_i\vert c_i- c^{'}_i\vert_1
   \end{aligned}
\end{equation}
where $o_i$ is the objectness score of the CLIP Proposal $C_i$ and $M$ is the number of the CLIP Proposal on one image.

\textbf{Cosine-similarity-based Classifier.}
By distilling the knowledge from CLIP, we are able to map the image feature of our model to CLIP feature space.
Instead of using a learnable linear layer as the classifier, we use text embedding generated from CLIP's text encoder. In the training phase, we only need the name of the base categories, which are then converted into text embedding $B_{i}$.
For each proposal $P_{i}$ given by the Region Proposal Network(RPN), we generate its feature $p_{i}$. The $p_{i}$ can be expressed as: $p_{i} = Conv_{c}(Align(Bbone(I), P_{i}))$, where $Conv_{c}$, $Align$, and $Bbone$ are the same as those in $c^{'}_{i}$ definition. The classification loss is given by:
\begin{equation}
   \label{eq:gradsimi}
   \begin{aligned}
   L_{cls} = \frac1N\sum_{i=1}^N L_{CE}(softmax(\boldsymbol{cos_{i}}) , y_{i})
   \end{aligned}
\end{equation}
where N is total number of proposals, $y_{i}$ is the assigned label for the proposal $P_{i}$.
The vector $\boldsymbol{cos_{i}}$ for the proposal $P_{i}$ is defined as $[cos(p_{i}, B_1),\dots, cos(p_{i}, B_n), cos(p_{i}, BG)]$ in which $n$ is number of base categories, $cos$ is the cosine similarity score, and $BG$ is a learnable vector for background.

At inference time, we also need to detect the novel categories. 
We generate the text embedding for both the base $B_{i}$ and the novel $N_{i}$.
The vector $\boldsymbol{cos_{i}}$ become $[cos(p_{i}, B_1), \dots, cos(p_{i}, B_n),cos(p_{i}, N_1), \dots, cos(p_{i}, N_k),$ $cos(p_{i}, BG)]$ where k is the number of the novel categories.

\textbf{Semantic-based Regressor.}
To improve the performance of the regression module, we add the semantic information of each category into consideration. We concatenate the text embedding with the proposal feature to predict the bbox.
For each foreground proposal $P_{i}$ given by the Region Proposal Network(RPN), we generate its feature for regression $r_{i}$. The $r_{i}$ can be expressed as $Conv_{r}(Align(Bbone(I), P_{i}))$, where $Align$ and $Bbone$ are the same as those in $c^{'}_{i}$ definition, and $Conv_{r}$ means passing the feature after the RoIAlign through the convolution layers and linear layers in the regression branch. The regression loss is defined as:
\begin{equation}
   \label{eq:gradsimi}
   \begin{aligned}
   L_{reg} = \frac1K\sum_{i=1}^K L_{1}(Linear(Cat(r_{i}, B_{y_{i}})) , a_{i})
   \end{aligned}
\end{equation}
where $K$ is the total number of the foreground proposals, $Linear$ is the linear layer for bbox prediction, $Cat$ mean concatenation, $B_{y_{i}}$ means the text embedding of the assigned GT label $y_{i}$ for the proposal $i$, and $a_{i}$ is the GT bbox for the proposal $P_{i}$. 
At inference time, since we no longer have the GT label. We concatenate the $r_{i}$ with the text embedding $B_{pred_{i}}$ or $N_{pred_{i}}$ of the predicted category of the proposal $P_{i}$, where $pred_{i} = \arg\max(\boldsymbol{cos_{i}})$.

Finally, our overall loss function is given by:
\begin{equation}
   \label{eq:gradsimi}
   \begin{aligned}
   L = L_{dist} + L_{cls} + L_{reg}
   \end{aligned}
\end{equation}

 \begin{table*}[!t]
    \centering
    \begin{tabular}{c|c|c|c|ccc}
      \hline 

       \hline
       \multirow{2}{*}{Method} & \multirow{2}{*}{Mask} & \multirow{2}{*}{Epoch} & \multirow{2}{*}{Novel Category Knowledge Source} & \multicolumn{3}{c}{AP50} \\ \cline{5-7} 
            & & & & Base & Novel & Overall \\ 
       \hline
       ZSD-YOLO\cite{xie2022zero}   &\xmark     & 50   & CLIP's feature                      & 31.7 & 13.6 & 19.0 \\
       CORA\cite{wu2023cora}       &\xmark     & 5    & CLIP                                    &35.5 & \textbf{35.1} & 35.4 \\
       F-VLM\cite{kuo2022f}      &\checkmark & 6    & CLIP                                    &- & 28.0 & 38.0 \\
       PBBL\cite{gao2021towards}       &\checkmark & -    & ALBEF\cite{li2021align}, CLIP, Pseudo-label  &46.1 & 30.8 & 42.1 \\
       OVOS\cite{Ma_2022_CVPR}       &\xmark     & 36   & CLIP's feature                       &51.3 & 20.3 & 43.2 \\        
       VTP-OVD\cite{long2022p}    &\checkmark & -    & CLIP's feature, Pseudo-label         & 31.5 & 29.8 & 46.1  \\
       OADP\cite{wang2023object}       &\checkmark &   -  & CLIP's feature, Pseudo-label            &53.3 & 30.0 & 47.2 \\
       CondHead\cite{wang2023learning}   &\checkmark & 107  & CLIP's feature                          &60.8 & 29.8 & \textbf{52.7} \\
       OV-DETR\cite{zang2022open}    &\xmark     & 50   & CLIP's feature                          &\textbf{61.0} & 29.4 & \textbf{52.7} \\

       ViLD\cite{gu2021open}       &\checkmark & 107  & CLIP's feature                       &59.5 & 27.6 & 51.3 \\ 
       \hline 
       Ours       &\checkmark & 36   & CLIP's feature, Novel Category Name & 59.9 & 31.6 &  52.1 \\ 
      \hline     
    \end{tabular}
    \centering
    \vskip -0.1in
    \caption{Evaluation results on COCO benchmark. All the models are trained with the ResNet-50 backbone. Mask indicates whether the model is trained with Mask annotations. Our model achieves 4\% improvement on Novel with 1/3 training time of ViLD.}
    \label{table:zeroshot}   
    \vskip -0.15in
 \end{table*}

\section{Experiments} \label{experiment}
We first present our model result on COCO\cite{lin2014microsoft} and LVIS\cite{gupta2019lvis} detection benchmark in section~\ref{zeroshot}.
In section~\ref{effiency} we compare our method with ViLD to show the efficiency of our method.
Finally, we conduct the ablation study with visualization analysis.

\textbf{Implementation Details.}
We use the publicly available pre-trained CLIP model ViT-B/32 as the open-vocabulary classification model, with an input size of 224$\times$224.

We finetune the LayerNorm layers in the CLIP with base categories instances in COCO or LVIS based on the setting and maintain all other parameters fixed.
All the instances are cropped by 1.2x enlarged GT bboxes.
We use an AdamW optimizer with a learning rate of 0.0001 and clip the L2 norm of the gradients when larger than 0.1. 
We finetune the model for 12 epochs.

For CLIP Proposal generation, we first resize the image with the image ratio maintained. 
We generate the anchors on each image with a stride of 32 pixels and with 5 different sizes (32, 64, 128, 256, 512), and 3 different ratios (1:1, 2:1, 1:2).
We select the top 1000 anchors after NMS as CLIP Proposals on each image.
If we will use the adapted CLIP features to train our detector we use the adapted CLIP to generate the CLIP Proposals.
Otherwise, we use the unadapted CLIP for CLIP Proposal generation.
In model training, we randomly select a fixed subset with 200 CLIP Proposals on each image for training.
We provide more implementation details in the supplementary material.

\subsection{Comparison with Current Methods} \label{zeroshot}
In this section, we evaluate EZAD in the COCO and LVIS detection benchmarks.

\textbf{Datasets and Evaluation Metrics.}
We evaluate EZAD on COCO and LVIS(v1). 
For the COCO dataset, we use train2017 for training and val2017 for validation. 
Following~\cite{zareian2021open}, we divide COCO into 48 base and 17 novel categories.
For the LVIS dataset, we use the training/validation images for training/evaluation.
Previous work uses Frequent and Common categories as the base (866 categories), and Rare categories as the novel (337 categories). 
We argue that in this split, the rare category objects are so sparse (less than 0.5\% annotations in the validation set) that the model's performance on it is not representative. 
Therefore, we propose a new split called LVIS-Fbase, which uses the Frequent categories as the base (405 categories), and both Common and Rare categories as the novel (common has 461 categories). 
On COCO, AP50 is used as the evaluation metric, while on LVIS the AP is used.

\textbf{Model.}
We train a Mask R-CNN\cite{he2017mask} model with ResNet-50\cite{he2016deep} FPN\cite{lin2017feature} backbone. The backbone is pre-trained on ImageNet\cite{deng2009imagenet}. We use SGD as the optimizer with batch size 4, learning rate 0.005, momentum 0.9, and weight decay 0.0001. We adopt linear warmup for the first 500 iterations, with a warm-up ratio is 0.001. 
On COCO, we train our model with 36 epochs and divide the learning rate by 10 at epoch 27 and epoch 33.
On LVIS, We train our model with 48 epochs and divide the learning rate by 10 at epoch 32 and epoch 44.
We train our model with multi-scale train-time augmentation.
\textbf{Baselines.}
We compare our method with different Zero-Shot Detection (ZSD) and Open-Vocabulary Detection (OVD) methods, which do not use additional datasets so that they can fit in the Zero-shot Annotation Detection (ZAD) setting.  
We mainly compare EZAD with ViLD\cite{gu2021open} and CondHead\cite{wang2023learning}, which uses distillation to obtain information on the novel categories from CLIP. 
They have the best overall performance on the COCO dataset. 
However, it uses the data augmentation of large-scale jittering\cite{ghiasi2021simple} with an extremely long training schedule.

\begin{table}[!t]
    \centering
    \begin{tabular}{c|c|cccc}
      \hline 

       \hline
       \multirow{2}{*}{Method} & \multirow{2}{*}{Epoch} & \multicolumn{4}{c}{AP} \\ \cline{3-6} 
            & & Freq & Comm & Rare & All \\ 
       \hline
       ViLD*      & 24   &24.9 & 12.2 & 11.2 & 17.5 \\
       Ours       & 24   &\textbf{30.9} & \textbf{14.3} & \textbf{12.5} & \textbf{20.5}\\ 
       ViLD*      & 48   &26.4          & 13.2 & 11.3 & 18.5\\ 
       Ours       & 48   &\textbf{31.9} & \textbf{15.2} & \textbf{13.1} & \textbf{21.3}\\ 
       
      \hline     
    \end{tabular}
    \centering
    \vskip -0.1in
    \caption{Evaluation results on LVIS-Fbase benchmark. The ViLD* is our reproduced result of ViLD with a shorter training schedule. EZAD outperforms ViLD in both 2x and 4x settings due to the efficient feature distillation.}
    \vskip -0.2in
    \label{table:lvis_efficiency}   
 \end{table}

\begin{table}[!t]
    \centering
    \begin{tabular}{c|c|cccc}
      \hline 

       \hline
       \multirow{2}{*}{Method} & \multirow{2}{*}{Epoch} & \multicolumn{3}{c}{AP50} \\ \cline{3-5} 
            & & Base & Novel & Overall \\ 
       \hline
       ViLD*      & 12   &48.3 & 17.2 & 40.2  \\
       Ours       & 12   &\textbf{55.7} & \textbf{30.4} & \textbf{49.0} \\ 
       ViLD*      & 36   &56.0 & 24.2 & 48.5\\ 
       Ours       & 36   &\textbf{59.9} & \textbf{31.6} & \textbf{52.1}\\ 
       
      \hline     
    \end{tabular}
    \centering
    \vskip -0.1in   
    \caption{Evaluation results on COCO. EZAD outperforms ViLD in both 1x and 3x settings, showing our distillation is more efficient.}
    \vskip -0.2in
    \label{table:coco_efficiency}  
 \end{table}

 \begin{table*}[!t]
    \centering
    \begin{tabular}{c|cccc|cccc|cccc}
      \hline 
      \multirow{2}{*}{Method} & \multicolumn{4}{c|}{Base} & \multicolumn{4}{c|}{Novel}  & \multicolumn{4}{c}{General} \\ \cline{2-13} 
                            & L             & M             & S    & Avg  & L    & M    & S    & Avg  & L    & M    & S    & Avg  \\ \hline 
       w/o Adaptation       & 69.9          & 70.2          & 46.6 & 66.8 & 90.7 & \textbf{82.3} & 48.5 & 77.7 & 61.3 & 62.2 & 36.9 & 53.9 \\
       w/ Adaptation        & \textbf{92.5} & \textbf{89.5} & \textbf{80.3} & \textbf{87.5} & \textbf{91.1} & 81.6 & \textbf{66.3} & \textbf{81.9} & \textbf{84.8} & \textbf{80.7} & \textbf{68.8} & \textbf{78.3} \\ 
      \hline     
    \end{tabular}
    \centering
    \vskip -0.1in  
    \caption{Adapting CLIP to the detection dataset's domain. The table presents the classification accuracy (ACC) of CLIP(w/ or w/o adaptation) when it is applied to classify the COCO dataset's instance. The ACC is aggregated based on the size of the instances. After the adaptation, the ACC is improved by a huge margin in three different settings, especially for small objects.}
    \vskip -0.15in
    \label{table:adapt_clip}   
 \end{table*}

\textbf{Results.}
Table~\ref{table:zeroshot} shows the results of EZAD in the COCO detection benchmark. 
Both the CORA\cite{wu2023cora} and F-VLM\cite{kuo2022f} are prompt-base methods. Although they need less training time and have strong novel category performance, their performance in the base categories is much worse than other methods.
ZSD-YOLO\cite{xie2022zero} and OVOS\cite{Ma_2022_CVPR} use a one-stage detector and suffer from high false positives in the detection results, which causes low AP in novel categories.
PBBL, VTP-OVD, and OADP are trained with pseudo-labels, introducing noise and harming the model performance on base categories.
OV-DETR\cite{zang2022open}, CondHead\cite{wang2023learning} and ViLD\cite{gu2021open} are pure distillation-based method. While the OV-DETR uses the transformer architecture which is heavier and has a better result than ViLD, the CondHead proposes a new detection head that can improve ViLD performance.
Our EZAD achieves 59.9\% and 31.6\% on base and novel, respectively. Its performance is 4\% better than ViLD in the novel and has a better performance in base and overall settings, with the use of 1/3 of ViLD's training time. 

Table~\ref{table:lvis_efficiency} shows the results of EZAD in our LVIS-Fbase benchmark.
The ViLD is trained 468 epochs on LVIS, which is too long to be fully reproduced.
We reproduce the ViLD with 24 and 48 epochs.
Our method uses ViLD's model structure and distillation pipeline, and the improvement of our method comes from the distillation feature and distillation region.
Therefore, reproducing the ViLD with a shorter training schedule will not affect the fairness of the comparison.
With 48 epochs, our method shows a 5\% improvement in Freq and 2\% improvement in both Comm and Rare over the ViLD with the same training schedule. 
In the CLIP proposals generation, EZAD only uses frequent and common object names to generate proposals. Therefore, EZAD does not take advantage of the category name on the Rare.
We believe that the enhancement in performance for the rare and frequent categories can be attributed to eliminating the domain gap, whereas the improvement in common categories results from both the CLIP proposals and domain adaptation. 

Because we have utilized the ZAD setting, there are novel instances present in our training images. This enables us to effectively leverage the information from these novel instances, thus achieving better performance than existing methods. However, the same cannot be achieved with Zero-Shot Detection (ZSD). On the other hand, we don't need additional annotations like many OVD methods do, meaning our ZAD setting can be applied to more real-life settings.

 \begin{table}[!t]
    \centering
    \begin{tabular}{c|c|ccc}
      \hline 

       \hline
       CLIP Feature & Distill Region & Base & Novel & Overall \\
       \hline
       Raw          & RPN Proposal  & 48.8 & 17.5 & 40.6 \\
       Adapted      & RPN Proposal  & 56.9 & 24.6 & 48.5 \\ 
       Raw          & CLIP Proposal & 48.7 & 19.3 & 41.7 \\ 
       Adapted      & CLIP Proposal & 55.7 &  \textbf{30.4} & 49.0 \\ 
      \hline     
    \end{tabular}
    \centering
    \vskip -0.1in    
    \caption{Ablation study on CLIP's feature adaptation and CLIP Proposal using COCO detection benchmark.}
    \vskip -0.1in
    \label{table:ablation_feat_and_clip_proposal}   
 \end{table}

 \begin{table}[!t]
    \centering
    \begin{tabular}{c|c|ccc}
      \hline 

       \hline
       SB Reg & PPDW & Base & Novel & Overall \\
       \hline
                   &            & 55.5 & 28.2 & 48.3 \\
        \checkmark &            & 55.8 & 29.8 & 48.5 \\ 
        \checkmark & \checkmark & 55.7 & \textbf{30.4} & 49.0 \\ 
      \hline     
    \end{tabular}
    \centering
    \vskip -0.1in    
    \caption{Ablation study on Semantic-based regressor and per CLIP Proposal distillation weight using COCO detection benchmark. SB Reg means the Semantic-based regressor, and PPDW means the per CLIP Proposal distillation weight.}
    \vskip -0.1in
    \label{table:albation_rwe_and_ppdw}   
    \vskip -0.1in
 \end{table}

 \begin{table}[!t]
    \centering
    \begin{tabular}{c|ccc}
      \hline 

       \hline
       Proposal & IoGT & \#(IoGT$\ge$0.8) & \#(IoGT$\ge$0.5) \\
       \hline
         RPN & 0.340 & 362818 (9\%)  & 610157 (15\%) \\
         CLIP(Ours) & \textbf{0.365} & \textbf{563799 (14\%)} & \textbf{870830 (21\%)} \\ 
         
      \hline     
    \end{tabular}
    \centering
    \vskip -0.1in  
    \caption{The effective distillation region of different proposals. The table presents the \textbf{I}ntersection between the proposal and the novel GT bboxes \textbf{o}ver novel \textbf{GT} bboxes (IoGT), the number of proposals that have high IoGT (IoGT$\ge$0.8, IoGT$\ge$0.5), and the percentage of these proposals in all proposals. Our CLIP proposals can cover more novel categories instances thus improving the distillation efficiency.}
    \vskip -0.15in 
    \label{table:iou_between_proposal_and_gtbbox}   
    \vskip -0.1in
 \end{table}

\subsection{Efficiency Evaluation} \label{effiency}
In this section, we compare EZAD with our reproduced ViLD to show the efficiency of our method. In Table~\ref{table:coco_efficiency}, we present our model and our reproduced ViLD on COCO with 1x and 3x training schedules. Our method is consistently better than ViLD in two different settings. Table \ref{table:lvis_efficiency} shows EZAD and our reproduced ViLD on the LVIS dataset with 2x and 4x training schedules. Our method shows substantial improvement over the ViLD with the same training schedule. 
These results suggest that the adapted feature space and the CLIP Proposals improve the distillation quality and efficiency. Thus, the model performance is improved.

\subsection{Ablation Study and Visualization}
In this section, we conduct ablation studies using the COCO detection benchmark. All the experiment details are the same as mentioned in section~\ref{zeroshot}. We train our detector for 12 epochs in all experiments of this section.

\begin{figure*}[!t]
   \centering
   \includegraphics[width=1\linewidth]{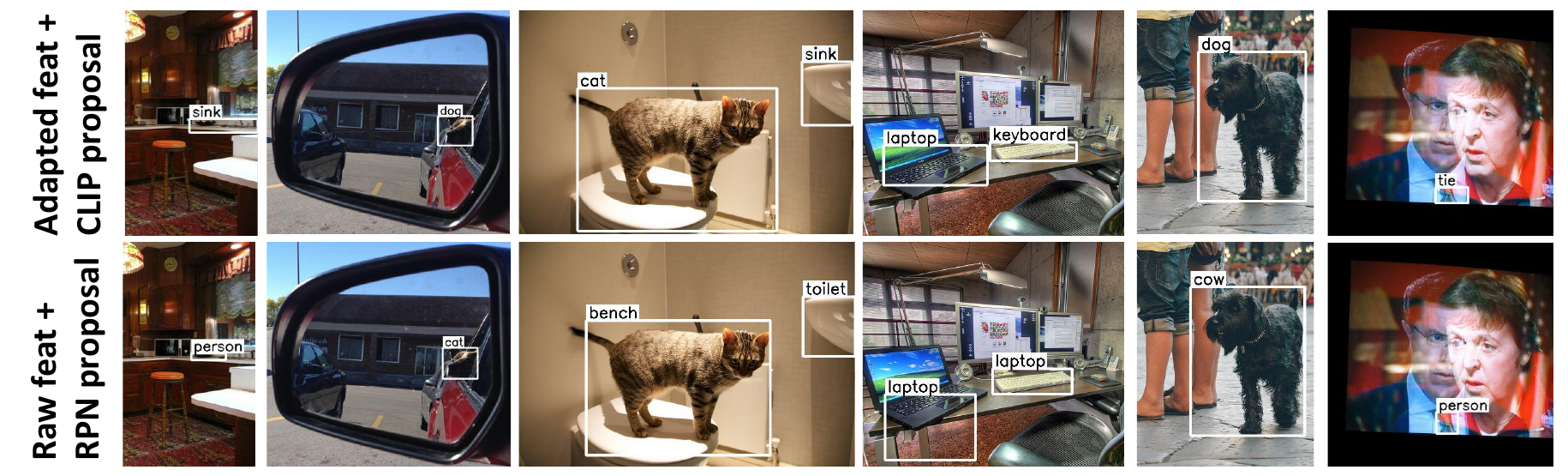} 
   \vskip -0.1in
   \caption{The visualization result on the COCO. The first row presents the results of the model trained with the adapted CLIP features from CLIP Proposals. The second row presents the results of the model trained with raw CLIP features from RPN proposals.} 
   \vskip -0.2in
   \label{detection_visualization} 
 \end{figure*}

\textbf{CLIP's Feature Adaptation and CLIP Proposal.} 
Table~\ref{table:adapt_clip} presents the classification result of adapting the CLIP to the COCO dataset's domain. 
We evaluate the classification accuracy(ACC) on the instances in the COCO validation set. 
We use the splits in Section~\ref{zeroshot}. 
Since the novel setting is a 17-ways classification which is easier than the other two settings, the ACC in the novel setting is much higher than the one in the other two.

Before the adaptation, the ACC in the general setting is only about 53.9\%, which is much lower than the ACC of the classifier in a well-trained detector. 
This phenomenon indicates that there is a huge domain gap between the training data of CLIP and the detection dataset.
While the objects in CLIP's training image are clear and large and at the center of the image, the objects in the detection dataset might be small and occluded.
Although we only fine-tuned the CLIP on base categories instances, we observed a huge improvement in all three settings, especially for the small objects.
The average ACC reaches 87.5\%, 81.9\%, and 78.3\% in the Base, Novel, and General settings, respectively. 
The ACC for the small objects improved by 33.7\%, 17.8\%, and 31.9\% in three different settings. 
This indicates that the simple fine-tuning method can effectively bridge the domain gap and make the feature more discriminating.

Table~\ref{table:ablation_feat_and_clip_proposal} shows the effectiveness of the CLIP Proposals and the CLIP's feature adaptation in the detection setting.
Using the adapted CLIP's feature for distillation can consistently improve both base and novel categories performance, no matter which kinds of distillation regions we use. 
Our model performance on novel benefits from using CLIP Proposals as distillation regions (30.4\% vs. 24.6\% with adapted CLIP features and 19.3\% vs 17.5\% with raw CLIP features).
We believe that the adaptation of the CLIP and CLIP Proposal are complementary to each other, which makes the improvement given by CLIP Proposals with adapted CLIP (5.8\%) larger than CLIP Proposals with raw CLIP (1.8\%).

The performance on base categories of models trained with CLIP Proposals is slightly worse than those trained with RPN proposals since the CLIP Proposals focus more on the regions with novel categories.
Our experiment results show that with a longer training schedule, the performance gap on the base categories can be eliminated while the advantage on the novel will be maintained. We will provide these results in the supplementary material.

\textbf{Semantic-based Regressor and Per CLIP Proposal Distillation Weight.}
Table~\ref{table:albation_rwe_and_ppdw} shows the effects of semantic-base regressor and per CLIP Proposal distillation weight. 
All experiments use adapted CLIP's features and CLIP Proposals as distillation regions.
The semantic-base regressor helps the model perform better on both base and novel categories, showing that the semantic meaning of the categories does provide useful information to the regressor.
Combining the semantic-base regressor and per CLIP Proposal distillation weight, the AP50 on novel reaches 30.4\%. 
This indicates the reweighting of different distillation boxes further improves the distillation quality.

\textbf{Statistic and Visualization Analysis.}
To compare which proposals can provide more meaningful novel categories information, we compare the effective distillation region of different proposals in an 8000 images subset of COCO training set in Table~\ref{table:iou_between_proposal_and_gtbbox}. 
We calculate the \textbf{I}ntersection between the proposal and the novel GT bboxes \textbf{o}ver novel \textbf{GT} bboxes (IoGT) and present the number and the percentage of the proposals that have high IoGT in all proposals.
Our CLIP proposals are 6\% higher than the RPN proposal in the percentage of the high IoGT proposals, meaning that the CLIP proposal can cover more potential novel instances and improve the distillation efficiency.

Figure~\ref{detection_visualization} provides some visualizations of the detected novel objects on the COCO benchmark. 
The first row presents the results from the model trained with the adapted CLIP features from CLIP Proposals (AFCP). 
The second row presents the results from the model trained with raw CLIP features from the RPN proposal (RFRP).
The results show that though two models can localize the object correctly, the AFCP model has a higher classification accuracy than the RFRP model thanks to the adapted features and the more meaningful distillation regions.

\section{Conclusion}
We propose a new setting for detecting novel instances, called Zero-shot Annotation object Detection (ZAD), to move one step closer to the real-life scenario.
Our method Efficient Feature Distillation for Zero-shot Annotation Detection (EZAD) in this paper. 
EZAD successfully bridging the domain gap between the classification and detection datasets, and selecting the meaningful distillation regions (CLIP Proposals) for obtaining knowledge from CLIP.
Benefiting from these two improvements, EZAD outperforms previous distillation-based work in both COCO and LVIS with a much shorter training schedule.
We believe our work provides a solid solution for applying zero-shot annotation detection in real life, and we hope our method can inspire other works in the future.

\section*{Acknowledgement}
This material is based on research sponsored by Air Force Research Laboratory (AFRL) under agreement number FA8750-19-1-1000. The U.S. Government is authorized to reproduce and distribute reprints for Government purposes notwithstanding any copyright notation therein. The
views and conclusions contained herein are those of the authors and should not be interpreted as necessarily representing the official policies or endorsements, either expressed or implied, of Air Force Laboratory, DARPA or the U.S. Government.
{\small
\bibliographystyle{ieee_fullname}
\bibliography{egbib}
}

\clearpage
\appendix
\section{Implementation Details}\label{sec:implementation details}
We include all implementation details in this section. 
\subsection{Adapt Image-language Model Feature} \label{sec:Adapt}
We use the publicly available pretrained CLIP\cite{radford2021learning} model ViT-B/32 as the open-vocabulary classification model, with an input size of 224$\times$224.

Based on the detection setting we use for training and evaluating our detector, we adapt the CLIP to two detection domains: COCO\cite{lin2014microsoft} detection domain, LVIS detection domain\cite{gupta2019lvis}.
We finetune the layer normalization layers in the CLIP with base category instances in COCO or LVIS based on the detection setting we use and maintain all other parameters fixed.
All base category instances are cropped by 1.2x enlarged GT bboxes.
We conduct the zero padding to convert each cropped region to the square and apply the default preprocessing pipeline of the CLIP.

We use CLIP to predict the category of each cropped region and calculate the cross-entropy loss with the GT label of each region.
We finetune the model by optimizing the Cross-Entropy Loss.
We use AdamW optimizer with a learning rate of 0.0001, batch size 4 and clip the L2 norm of the gradients when larger than 0.1. 
We finetune the model for 12 epochs.

\subsection{Generate CLIP Proposals}
When generating the CLIP Proposals, we still use the CLIP model we mentioned in section~\ref{sec:Adapt} as a classifier to select the distillation regions. If we will use the adapted CLIP's feature to train the detector, we will use the adapted CLIP to generate the CLIP Proposals. Otherwise, we use the unadapted CLIP to generate CLIP Proposals.

We generate the CLIP proposals on all the training images of the detection dataset base on the detection setting we use.
We first resize the image with the image ratio maintained. 
The long edge of the image will be resized into 1333 as width or 800 as height.

We generate the anchors on each image with a stride of 32 pixels and with 5 different sizes (32, 64, 128, 256, 512), and 3 different ratios (1:1, 2:1, 1:2).
We select the top 1000 anchors after NMS as CLIP Proposals on each image.
We filter out the anchors which have high IoU with the base category GT bboxes to reduce the redundancy since we will add 1.2x enlarged base category GT bbox as part of the CLIP Proposals.
In model training, we randomly select a fixed subset with 200 CLIP Proposals on each image for training.

\subsection{Detection Setting}
In COCO detection setting, the dataset is divided into 48 base categories and 17 novel categories.
15 categories without a synset in the WordNet hierarchy are removed. 

We filter out the training images which do not have base category annotation.
Following the setting in \cite{zareian2021open}, we filter out the images that have neither the base category instances nor the novel category instances in the validation set. 
The training set contains 107761 images and 665387 base category instances.
The validation set contains 4836 images and 28538 base category instances and 33152 novel category instances.
We evaluate the model in a generalized setting, which evaluates the base and novel categories at the same time.
AP50 is used as the evaluation metric.

In LVIS detection setting, the dataset is divided into 866 base categories (containing 405 frequent categories and 461 common categories) and 337 novel categories (337 rare categories).
Our LVIS-Fbase split uses the frequent categories as the base(405 categories), common and rare categories as the novel(common has 461 categories, rare has 405 categories).
The training set contains 98531 images and 1200258 base category instances.
The validation set contains 19442 images and 230427 base category instances and 14280 novel category instances.
We aggregate the model performance in frequent, common, and rare categories separately.
AP is used as the evaluation metric.

\section{Experiments in Few-shot Detection Settings}\label{sec:fewshot}

In few-shot object detection, the model is trained on the base category's annotations and evaluated on novel categories. 
The only difference is that in few-shot detection, each novel category has the same number of annotated objects(i.e, K-shot), which can be used to improve the model performance on the novel before the model is evaluated.
We directly evaluate our model in the few-shot benchmark, without using this K-shot additional information.

\textbf{Datasets and Evaluation Metrics.} 
We evaluate our approach on PASCAL VOC 2007+2012 and COCO.
For the few-shot PASCAL VOC dataset, we combine the trainval set of 2007 with the one of 2012 as training data. 
PASCAL VOC 2007 test set is used for evaluation. 
The 20 classes are divided into 15 base classes and 5 novel classes. 
We evaluate our model in three different base/novel splits used in \cite{wang2020frustratingly}.
Split 1 has 14631 training images with 41084 base category instances, and the validation set has 4952 images, 10552 base category instances, and 1480 novel instances.
Split 2 has 14779 training images with 40397 base category instances, and the validation set has 4952 images, 10447 base category instances, and 1585 novel instances.
Split 3 has 14318 training images with 40511 base category instances, and the validation set has 4952 images, 10605 base category instances, and 1427 novel instances.

For the few-shot COCO dataset, we use the COCO train2017 as training data and evaluate our model on the COCO val2017.
The 20 categories that exist in PASCAL VOC are used as the novel categories, while the rest of the 60 categories are used as the base categories.
The training set has 98459 images and 367189 base category instances.
The validation set has 5000 images and 15831 base category instances and 36781 novel category instances.

AP50 is used as the evaluation metric in PASCAL VOC, while AP and AP50 are used in COCO.

\textbf{Model.}
Following previous work in few-shot detection, we train a Faster R-CNN\cite{ren2015faster} model with ResNet-101 FPN backbone. 
The backbone is pretrained on ImageNet. 
We use SGD as the optimizer with batch size 4, learning rate 0.005, momentum 0.9, and weight decay 0.0001. 
We also adopt linear warmup for the first 500 iterations, with a warm up ratio is 0.001. 
We apply multi-scale train-time augmentation.
For the PASCAL VOC dataset, we train the model for 21 epochs and divide the learning rate by 10 at epoch 15 and epoch 18.
For the COCO dataset, we train the model for 18 epochs and divide the learning rate by 10 at epoch 14 and epoch 16.

\textbf{Baselines.}
We compare EZAD's performance with two few-shot detection models, TFA\cite{wang2020frustratingly} and Meta Faster R-CNN \cite{han2022meta} as the baselines. The TFA model with linear layer as the classifier is noted as \textit{TFA w/fc}, while the model with cosine classifier is noted as \textit{TFA w/cos}.

\textbf{Results.}
Table \ref{table:voc_fewshot} shows the results on the PASCAL dataset. 
EZAD achieves 40.9\% in novel AP50 averaged over three different splits.
EZAD's performance matches the TFA 3-shot performance in split1 and split2 and is 4.7\% higher than TFA in split3. 
Compared with the TFA's performance on base, EZAD is 1.8\% higher.
For Meta Faster R-CNN, it generates proposals for each category on each image, which needs multiple forward passes. 
Its inference time will be much slower if the dataset has a large number of novel categories.
Compared with the Meta Faster R-CNN, EZAD outperforms it without using any additional annotations by a 1.6\%, 3\%, and 6.9\% in three different splits, respectively.
Table \ref{table:coco_fewshot} shows the results on the COCO dataset.
EZAD achieves 10.2\% and 22.2\% in AP and AP50, respectively, matching TFA's 10-shot performance and 2.6\% and 5.9\% higher than the Meta Faster R-CNN's 2-shot performance in AP and AP50, respectively.
Our model zero-shot performance on the few-shot setting shows the power of adapted multi-modal feature space and validates the effectiveness of using CLIP Proposals as distillation regions.

 \begin{table}[!t]
    \centering
    \begin{tabular}{c|c|cccc}
      \hline 

       \hline
       \multirow{2}{*}{Method} & \multirow{2}{*}{Shot} & \multicolumn{4}{c}{Novel AP50}  \\ \cline{3-6} 
       
        & &Split1         & Split2        & Split 3       & Avg \\
       \hline
       TFA w/fc          & 1    & 36.8          & 18.2          & 27.7          & 27.6 \\
       TFA w/fc          & 2    & 29.1          & 29.0          & 33.6          & 30.6 \\ 
       TFA w/fc          & 3    & 43.6          & 33.4          & 42.5          & 39.8 \\ 
       TFA w/cos         & 1    & 39.8          & 23.5          & 30.8          & 31.4 \\
       TFA w/cos         & 2    & 36.1          & 26.9          & 34.8          & 32.6 \\ 
       TFA w/cos         & 3    & \textbf{44.7} & \textbf{34.1} & 42.8          & 40.5 \\ 
       MF R-CNN          & 1    & 43.0          & 27.7          & 40.6          & 37.1\\
       \hline 
       Ours              & 0    & 44.6          & 30.7          & \textbf{47.5} & \textbf{40.9} \\ 
      \hline  
       \multicolumn{6}{c}{\textbf{Split1 Base(AP50): TFA (3-Shot)=79.1, Ours=80.8}} \\
       \hline
    \end{tabular}
    \centering
    \vspace{+1mm}    
    \caption{Evaluation results on the novel categories of PASCAL VOC few-shot benchmark. MF R-CNN means Meta Faster R-CNN. Our model zero-shot performance on the novel match the TFA's performance in its 3-shot setting. Our model also has a better performance on base.}
    \label{table:voc_fewshot}   
 \end{table}
 
 \begin{table}[!t]
    \centering
    \begin{tabular}{c|c|c c}
      \hline 

       \hline
       Method            & Shot & AP & AP50        \\
       \hline
       TFA w/fc          & 10    & 10.0  & 19.2      \\
       TFA w/cos         & 10    & 10.0  & 19.1      \\ 
       MF R-CNN          & 2     & 7.6   & 16.3     \\
       \hline 
       Ours              & 0     & \textbf{11.0}  & \textbf{23.5}         \\ 
      \hline  
    \end{tabular}
    \centering
    \vspace{+1mm}    
    \caption{Evaluation results on novel categories of COCO few-shot benchmark. MF R-CNN means Meta Faster R-CNN. Our model zero-shot performance on the novel match the TFA's performance in its 10-shot setting.}
    \label{table:coco_fewshot}   
 \end{table}

\section{Additional Ablation Study}\label{sec:ablation}
Table \ref{table:bbox_size} presents the experimental results on how the size of the bounding box (bbox) that we use to crop the instances in the COCO~\cite{lin2014microsoft} dataset affects the classification accuracy (ACC) of the unadapted CLIP~\cite{radford2021learning}.
For the large objects, the more accurate bbox provided the higher ACC CLIP can achieve. For the small objects, CLIP needs more background information to be correctly classified. In all settings, the average ACC over all three sizes of the bbox is still much lower than the classifier of the well-trained detector, indicating the domain gap between the training data of the CLIP and the detection dataset exists. We use the 1.2x GT bbox to crop the base GT instance since it has the highest average ACC.

We provide an additional ablation study in Table \ref{table:ablation_feat_and_clip_proposal_36e}. We train all models with the adapted CLIP features. For the models trained with 12 epochs, the performance on novel categories of the model trained with the RPN proposals is 5.8\% lower than the one of the model trained with the CLIP proposals, though the former has slightly better performance on base categories. For the models trained with 36 epochs, two models (RPN proposal and CLIP proposal) has similar performance on base categories, and the model trained with the CLIP proposal features still have much better novel category performance. This indicates that the negative effect on model performance on base categories caused by the CLIP proposal is negligible and can be alleviated by a longer training schedule.
It also shows that the information of base categories provided by the distillation has redundancy, which may accelerate the model convergence on base, but may not improve the model performance.

 \begin{table}[!t]
    \centering
    \begin{tabular}{c|cccc}
      \hline 
      \multirow{2}{*}{Bbox Size} & \multicolumn{4}{c}{General} \\ \cline{2-5} 
                            & L    & M    & S    & Avg  \\ \hline 
       0.8x GT       & 62.3          & 54.0          & 23.2          & 47.1 \\
       1.0x GT       & \textbf{64.0} & 61.9          & 32.9          & 53.4 \\
       1.2x GT       & 61.3          & \textbf{62.2} & 36.9          & \textbf{53.9} \\
       1.5x GT       & 56.7          & 59.5          & 40.6          & 52.6 \\
       2.0x GT       & 50.5          & 52.6          & \textbf{42.9} & 48.9 \\
      \hline     
    \end{tabular}
    \centering
    \vspace{+1mm}    
    \caption{The classification accuracy (ACC) of the unadapted CLIP on COCO instances with different sizes of the GT bboxes to crop the instances. We decide to use the 1.2x enlarged GT bbox to crop the instance since it has the best average ACC.}
    \label{table:bbox_size}   
 \end{table}

 \begin{table}[!t]
    \centering
    \begin{tabular}{c|c|ccc}
      \hline 

       \hline
       Epoch & Distill Region & Base & Novel & Overall \\
       \hline

       12      & RPN Proposal  & 56.9          & 24.6          & 48.5 \\ 
       12      & CLIP Proposal & 55.7          & 30.4 & 49.0 \\ 
       36      & RPN Proposal  & \textbf{60.2} & 24.3          & 50.8 \\ 
       36      & CLIP Proposal & 59.9          & \textbf{31.6} & \textbf{52.1} \\ 
      \hline     
    \end{tabular}
    \centering
    \vspace{+1mm}    
    \caption{Ablation study on using CLIP Proposals as distillation in COCO benchmark. The model trained with CLIP Proposals has much better performance on novel categories.}
    \label{table:ablation_feat_and_clip_proposal_36e}   
 \end{table}

\section{Additional Visualizations}\label{sec:visualization}

\begin{figure*}[!t]
   \centering
   \includegraphics[width=1\linewidth]{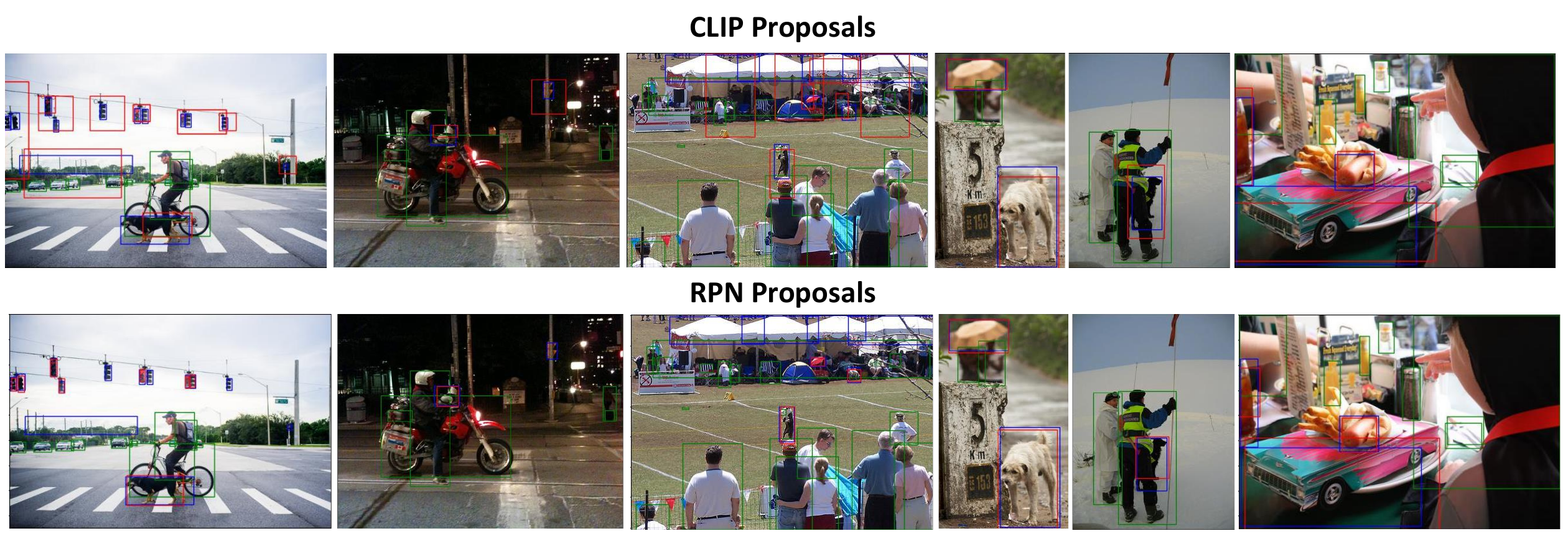} 
   \caption{Visualization of using CLIP Proposals or RPN proposals as distillation regions in COCO setting. The blue boxes and green boxes represent the GT bboxes of the novel and base categories. The red boxes represent the CLIP proposals or the RPN proposals with the highest IoU with the novel GT bboxes. The visualization shows the CLIP proposals can cover more novel objects even though the box may not accurate.} 
   \label{clip_proposal_visualization} 
   \vskip -0.1in
 \end{figure*}
 
Fig~\ref{clip_proposal_visualization} shows the visualization of using CLIP Proposals and RPN proposals as distillation regions in the COCO setting. 
The blue boxes and green boxes represent the GT bboxes of the novel and base categories. 
The red boxes represent the CLIP Proposals or the RPN proposals with the highest IoU with the novel GT bboxes. 
The three images on the left show that the CLIP Proposals can cover most of the novel category objects although the boxes may not accurate, while the RPN regards some of the novel objects as background and just ignores them. 
Although the CLIP proposals are not accurate, the features extracted from these boxes are accurate and meaningful. 
This phenomenon is also proved by the experiments in \cite{gu2021open}.
Therefore, using the CLIP Proposals as distillation regions provides more novel category information and improve the detector's performance on the novel.

\begin{figure*}[!t]
   \centering
   \includegraphics[width=1\linewidth]{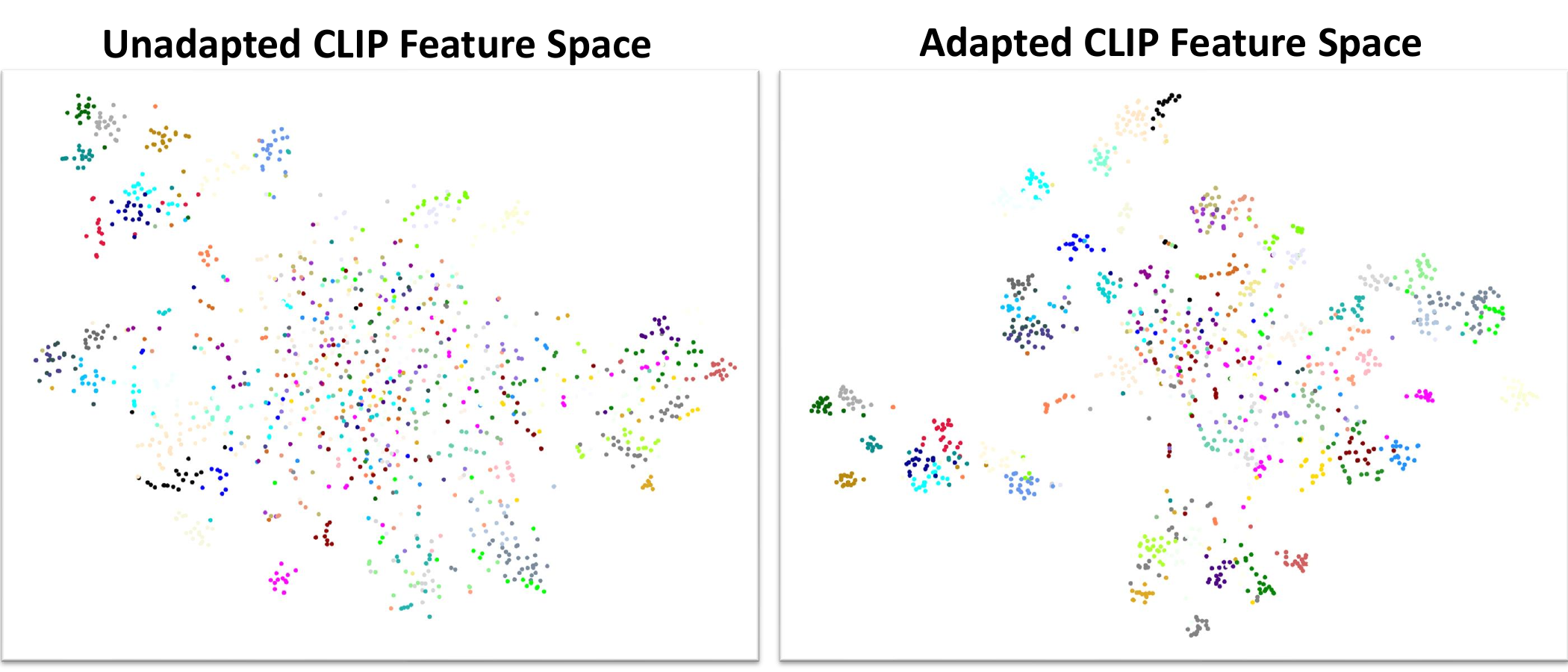} 
   \caption{The tSNE embeddings of the COCO GT instance feature from the unadapted CLIP and adapted CLIP. The GT features from the adapted CLIP form more dense clusters, indicating that the features become more discriminating and the CLIP is adapted into the detection dataset domain.} 
   \label{tsne_map} 
   \vskip -0.1in
 \end{figure*} 

Fig~\ref{tsne_map} shows the tSNE embeddings of the COCO instance features of the unadapted CLIP and the adapted CLIP. 
We collect 20 GT instances for each base and novel category in COCO setting and extract their features from unadapted CLIP or adapted CLIP, and then generate the tSNE embeddings with these features.
The GT instances in the adapted CLIP feature space form some dense clusters. This indicates that the CLIP's feature space has been adapted in the COCO dataset domain and the features become more discriminating after adaptation, improving the classification accuracy. 
The dots do not form a dense cluster mostly come from the "person" category. Since the instances of the person usually show up with other categories instances and occluded by other objects, therefore the person categories features are more scattered.

\end{document}